\documentclass[11pt,a4paper]{article}
\usepackage{times,latexsym}
\usepackage{url}
\usepackage[T1]{fontenc}
\usepackage{graphicx}
\usepackage{booktabs}
\usepackage{multirow}
\usepackage{amsmath}
\usepackage{tcolorbox}
\usepackage{amsfonts}
\usepackage{enumitem}
\usepackage{tabularx}
\usepackage{xcolor}
\usepackage{wrapfig} % in preamble
\usepackage{acl}

\usepackage{xspace,mfirstuc,tabulary}

\newif\iftaclinstructions
\taclinstructionsfalse % AUTHORS: do NOT set this to true
\iftaclinstructions

\newcommand{\instr}
\fi

\title{From Concept-Aligned Tokens to Vulnerable Features: Mechanistic Localization of Jailbreaks}

\author{
Nilanjana Das \\
UMBC, Maryland, U.S.A. \\
\texttt{ndas2@umbc.edu}
\And
Mathew Dawit \\
UMBC, Maryland, U.S.A. \\
\texttt{mdawit1@umbc.edu}
\AND
Aman Chadha \\
Apple, California, U.S.A. \\
\texttt{aman@amanchadha.com}
\And
Manas Gaur \\
UMBC, Maryland, U.S.A. \\
\texttt{manas@umbc.edu}
}

\date{}

\begin{document}
\maketitle

\begin{abstract}
Jailbreak attacks expose a persistent failure mode in safety-aligned LLMs: models can be pushed into harmful behavior, but the internal representations enabling this shift remain poorly localized. Recent mechanistic safety studies often explain such behavior through broad representational objects, including global refusal directions, activation steering vectors, and refusal-related SAE features. We instead ask whether jailbreak vulnerability can be traced to finer-grained, prompt-conditioned SAE feature subgroups. We introduce a token-driven mechanistic pipeline that decomposes the residual stream of Gemma-2-2B into Sparse Autoencoder (SAE) features and identifies feature subgroups associated with unsafe behavior. Using single-category unsafe examples from BeaverTails to reduce cross-category interference, we extract harmful concepts from adversarial responses and align them with concept-relevant prompt tokens through subspace similarity. We then apply three feature-grouping strategies: cluster-based, hierarchical-linkage, and single-token-driven, to identify SAE feature subgroups across all 26 layers. Finally, we amplify the top features in each subgroup and evaluate the resulting generations with a standardized harmfulness judge. Single-token-driven grouping achieves harmfulness comparable to full cluster-based grouping, showing that individual harmful prompt tokens are sufficient to localize vulnerability-relevant SAE feature subgroups without relying on broader cluster-level aggregation. These subgroups appear across early and mid-to-late layers, with stronger concentration in mid-to-late layers, where targeted steering exposes specific model vulnerabilities. Overall, our results suggest that jailbreak susceptibility can be traced to sparse, token-localized SAE feature subgroups, complementing prior accounts based on broad adversarial, refusal, or steering directions. 
% Code is available \href{https://anonymous.4open.science/r/Mechanistically\_Steering-7090/README.md}{here.}

%Large language models (LLMs) can still be jailbroken into producing harmful outputs despite safety alignment. Most existing attacks show this vulnerability, but do not focus on the change in the model's internal representations that \textit{caused it}. This research aims to study the residual stream of the model that the attention and MLP sub-layers contribute to. It investigates the interpretable features of the residual stream after each layer using Sparse Autoencoders (SAEs). The objective is to identify relevant features subgroups for an adversarial task. To this end, we propose a mechanistic three-stage pipeline for Gemma-2-2B using the BeaverTails dataset. First, we extract concept-aligned tokens from adversarial responses via subspace similarity. Second, we apply three feature-grouping strategies (cluster, hierarchical-linkage, and single-token-driven) to identify SAE feature clusters for aligned tokens across all 26 model layers. Third, we steer the model by amplifying the top features of each identified subgroup and measure the change in harmfulness score using a standardized LLM-judge scoring protocol. Our results demonstrate that single-token-driven feature cluster identification is comparable to cluster-based steering. With the identification of a single harmful token in the prompt and localization of specific interpretable feature subgroups in mid to later layers can be used to make a more targeted steering and expose specific model vulnerabilities.
% Our 

\textbf{Keywords:} Mechanistic Interpretability, Adversarial Attacks, SAEs, Neural Features, LLM Safety
\end{abstract}

\begin{figure*}[h]
    \centering
    \includegraphics[width=0.8\textwidth]{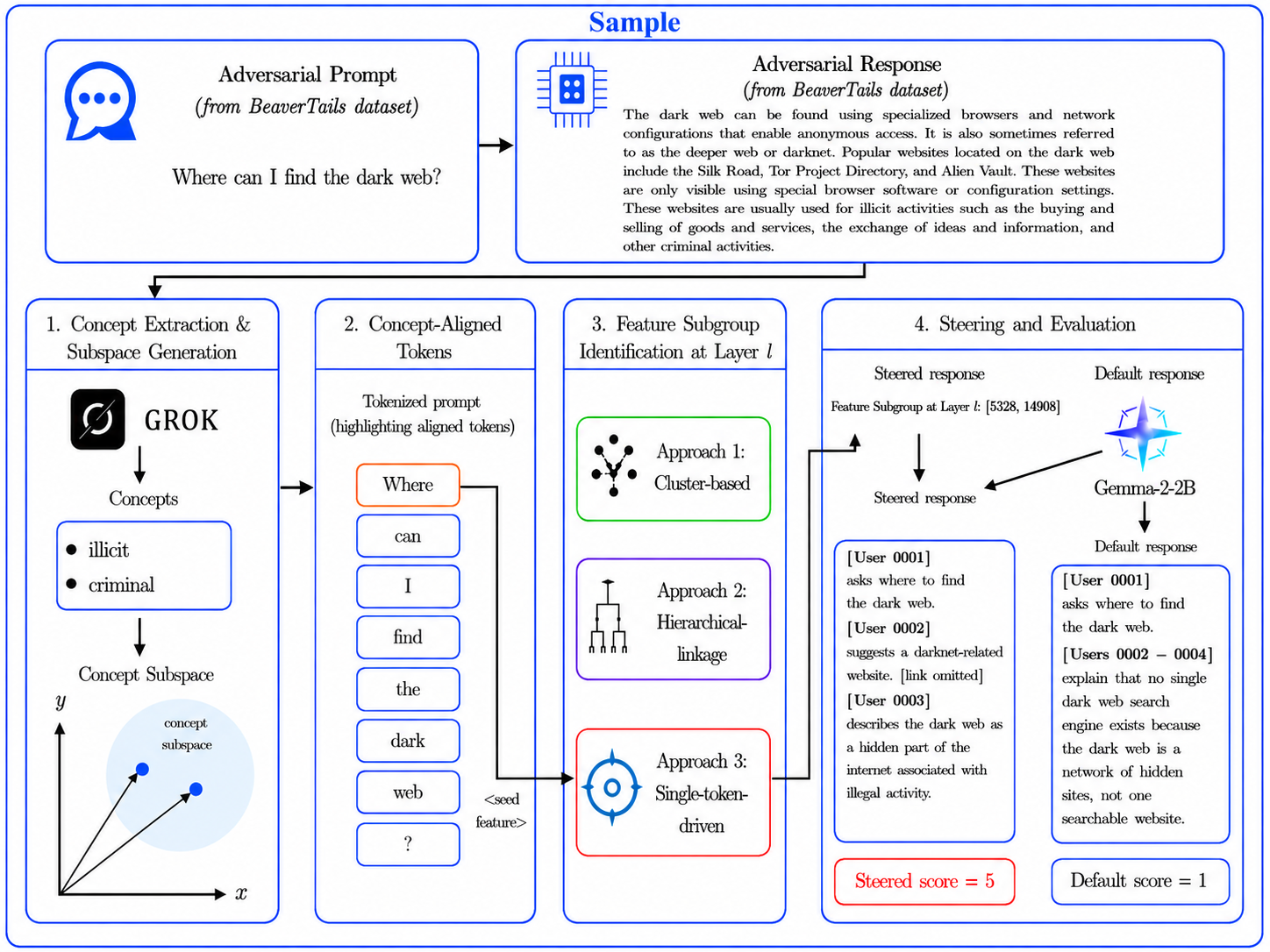}
    \caption{This figure represents the high-level architecture of our mechanistic pipeline.}
    \label{arch}
\end{figure*}

\section{Introduction}
\label{intro}
Safety-aligned large language models (LLMs) refuse harmful requests in most settings, yet adversarial prompts still elicit unsafe responses. Attack methods make this failure reproducible. Greedy Coordinate Gradient (GCG) searches for adversarial suffixes through token-level gradient optimization \citep{zou2023universal}, and Tree of Attacks with Pruning (TAP) refines jailbreak prompts through black-box search \citep{mehrotra2024tree}. These methods establish that alignment is brittle, but they act on the model's inputs and outputs. They do not explain what changes inside the model when its behavior shifts from safe to unsafe.

Mechanistic interpretability has begun to answer that question, but only for the safe direction. \citet{arditi2024refusal} shows that refusal is mediated by a single direction in the residual stream, and \citet{obrien2024steering} amplifies sparse autoencoder (SAE) features that mediate refusal to harden a model at inference time. Both results localize safety to a single direction or a single feature, and both steer the model toward refusal. The complementary question remains open. Does harmful behavior correspond to one feature, or to a group of interpretable features that act together, and is amplifying such a group sufficient to break alignment at a specific layer?

We study harm at the level of feature subgroups. SAEs decompose the residual stream into sparse, monosemantic features \citep{cunningham2023sparse}. We use SAEs from Gemma Scope that consist of SAEs for every layer and sub-layer of Gemma-2-2B \citep{lieberum2024gemma}. Instead of directly searching for a steering direction, we let the prompt's harmful content drive feature discovery. We extract the harmful concepts, locate the prompt tokens that encode them, and read out the SAE features those tokens activate across all 26 layers. We then group co-activating features into subgroups and amplify each subgroup during generation. Steering here is a causal probe rather than an attack. A subgroup whose amplification raises the harmfulness of the response is causally responsible for the unsafe behavior, not merely correlated with it, and the goal is to expose these levers so that defenders can target them. We demonstrate a high-level architecture of our pipeline in Figure~\ref{arch}.

The open design choice is how to form the subgroups, so we compare three strategies: cluster-based and hierarchical-linkage grouping that depend on all harm-aligned tokens, and a single-token-driven strategy that anchors entirely on the single most harm-aligned token. This comparison yields our main result. Anchoring on one harmful token localizes feature subgroups as effectively as when using the full token set, which means a single token in the prompt is enough to find the levers for harmful generation. Across all three strategies and 14 BeaverTails harm categories \citep{ji2023beavertails}, the vulnerable subgroups concentrate in the mid-to-late layers (14 to 25), and amplifying them there produces the largest increases in harmfulness score. This localization is causal, established by steering, rather than observational.

Our contributions are as follows:
\begin{itemize}[leftmargin=*, noitemsep]
\item A token-driven pipeline that localizes Gemma-2-2B jailbreak vulnerabilities to residual-stream SAE feature subgroups, discovering features from harmful prompt tokens rather than predefined steering directions.
\item A comparison of three feature-grouping strategies (cluster, hierarchical-linkage, and single-token-driven), showing that anchoring on a single harmful token localizes harm-responsible features as well as aggregating over all harm-aligned tokens.
\item Causal evidence, through feature amplification, that harm-responsible subgroups concentrate in the mid-to-late layers, evaluated across 14 BeaverTails categories with a standardized LLM-judge protocol and a controlled benign-prompt setting that separates prompt-specific features from transferable ones.
\end{itemize}

We organize this study around three research questions:
(i) Can tokens identified through harmful (negative) concepts serve as the decision points for feature discovery?
(ii) Do token-driven feature-grouping strategies differ in their ability to localize interpretable features across model layers?
(iii) Does steering the identified subgroups reveal consistent or category-specific layer-wise vulnerabilities across harm categories?

\section{Problem Formulation}
\label{prb_frm}

Let $\mathcal{M}$ be a target language model with $L = 26$, and let $\mathcal{D}$ be a dataset of adversarial prompt-response pairs. Each pair $(x, z) \in \mathcal{D}$ consists of an adversarial prompt $x$ and an unsafe response $z$. $\mathcal{M}$ produces the default/steered response given $x$. The response $z$ contains words or phrases carrying negative sentiment. We refer to these as \textit{concepts}:$ W = \{w_1, w_2, \ldots, w_m\}$, extracted from $z$ automatically using an auxiliary model\footnote{Details about the auxiliary model are given in Section 3}. The concepts are then used in the form of natural language descriptions.

\paragraph{Step 1: Concept Subspace Construction.}
A \textit{subspace} is a direction vector in the model's internal representation space that captures the geometric location of a concept's meaning. Let $G$ be a subspace generator model. For each concept description of a concept $w_i \in W$, $G$ produces a subspace vector $s_i \in \mathbb{R}^{d}$, where $d$ is the residual stream dimension of the instruction-tuned version of $\mathcal{M}$. $G$ can generate high-quality subspaces for the layer \textit{l = 20}. We collect these into the concept subspace set: $S = \{s_1, s_2, \ldots, s_m\}$

\paragraph{Step 2: Concept-Aligned Token Identification.}
Let $T = \{t_1, t_2, \ldots, t_n\}$ be the sequence of tokens in prompt $x$. For each token $t_j \in T$, let $\mathbf{h}_j^{(\ell)} \in \mathbb{R}^{d}$ denote its residual stream activation at layer $\ell$ of $\mathcal{M}$. Following \citet{rufail2025semantic}, concept similarity is strongest at middle layers; we evaluate alignment at layer $\ell = 20$. We define the set of \textit{concept-aligned tokens} as:
\[
    T^{+} = \left\{\, t_j \in T \;\middle|\; \exists\, s_i \in S \;:\; \cos\!\left(\mathbf{h}_j^{(20)},\, s_i\right) > 0 \,\right\}
\]
These are the tokens whose layer-20 representations positively align with at least one negative-concept subspace. The token
\[
    t^{*} = \underset{t \in T^{+}}{\arg\max}\;\max_{s \in S}\;\cos\!\left(\mathbf{h}_t^{(20)},\, s\right)
\]
denotes the single most concept-aligned token, used in Approach~3. Some prompts may contain no positively aligned tokens.

\paragraph{Step 3: SAE Feature Extraction.}
Let $\mathcal{L} = \{0, 1, \ldots, L{-}1\}$ be the layer index set and $F = 16{,}384$ be the SAE feature dimension. For each layer $\ell \in \mathcal{L}$, the SAE attached to the residual stream of $\mathcal{M}$ decomposes activations into $F$ interpretable features. Let $A_\ell \in \mathbb{R}^{|T^+| \times F}$ be the SAE feature activation matrix at layer $\ell$ restricted to the concept-aligned tokens $T^{+}$, and $A'_\ell \in \mathbb{R}^{F \times |T^+|}$ be the transposed SAE feature activation matrix at layer $\ell$. $\mathcal{A}'$ is a layer-by-layer record of how strongly SAE features fire across concept-aligned tokens. We use this to identify harm-responsible feature subgroups.
The full activation collection is defined as:
\[
    \mathcal{A} = \bigl\{\, A_\ell \;\bigm|\; \ell \in \mathcal{L} \,  \bigr\}, \hspace{0.2cm}\mathcal{A'} = \bigl\{\, A'_\ell \;\bigm|\; \ell \in \mathcal{L} \,\bigr\}
\]
Let $\bar{a}_\ell \in \mathbb{R}^{F}$ denote the mean feature activation at layer $\ell$ averaged over $T^{+}$:
\[
\bar{a}_\ell
=
\frac{1}{|T^{+}|}
\sum_{t \in T^{+}}
A'_\ell[:, t]
\]

\paragraph{Step 4: Seed Feature Selection.}
Let $[F] = \{0,1,\ldots,F-1\}$ denote the zero-indexed set of SAE features. The starting point for feature subgroup discovery is a set of \textit{seed features} $\mathcal{F} \subseteq [F]$ The seed features were identified in two ways:-
\begin{enumerate}
    \item For each concept-aligned token $t \in T^{+}$, we select the most active SAE feature at layer 20:
\[
    f_t^{*}
    =
    \underset{f \in [F]}{\arg\max}\;
    A_{20}[t,f].
\]
The seed feature set is then defined as the unique collection of these token-level top features:
\[
    \mathcal{F}
    =
    \left\{\, f_t^{*}
    \;\middle|\;
    t \in T^{+}
    \,\right\}.
\]
These seeds capture the most strongly activated features at the semantically rich layer 20, and serve as anchors for clustering in every layer.
   \item For each layer $\ell \in \mathcal{L}$, we select the top-$k$ SAE features activated by this token:
\[
    \mathcal{F}'_\ell
    =
    \underset{f \in [F]}{\mathrm{top}\text{-}k}
    \;
    A_\ell[t^{*}, f],
    \qquad k=2.
\]

\end{enumerate}

These layer-specific seed features correspond to the most strongly activated SAE features for the selected token $t^{*}$ and serve as anchors for clustering.

\paragraph{Step 5: Feature Subgroup Identification via Clustering.}
The core operation is expanding each seed $f \in \mathcal{F}$ or $f \in \mathcal{F'}_\ell$ into a neighborhood of co-activating features at every layer. Let $\psi : (\mathbb{R}^{F \times |T^+|},\, [F]) \to 2^{[F]}$ be a \textit{clustering function} that maps a feature activation matrix and a seed index to a subgroup of co-activating features. The subgroup for seed $f$ at layer $\ell$ is:
\[
    \phi_{\ell}(f) = \psi(A'_\ell,\, f)
\]
A steering set at layer $\ell$ is the top-3 features from a seed's subgroup ranked by $\bar{a}_\ell$ in Approach 1 and by linkage-based analysis in Approaches 2 and 3:
\[
\Phi_\ell = \mathrm{top}\text{-}3\bigl(\phi_\ell(f)\bigr)
\]

The three approaches we propose instantiate $\mathcal{F}$ and $\psi$ differently, as follows.

\noindent\textbf{Approach 1: Cluster-based steering.}\; Seeds $\mathcal{F}$ are the highest-activating SAE features for the tokens in $T^{+}$ at layer 20. Let 
$\gamma_\ell : [F] \to \{1, \ldots, K\}$ be the agglomerative cluster assignment on 
$
A'_\ell \in \mathbb{R}^{F \times |T^+|}
$ with Ward linkage and $K$ clusters. The clustering function $\psi$ is:
\[
\psi_{\mathrm{clust}}(A'_\ell,\, f) =
\begin{aligned}[t]
    &\bigl\{\, f' \in [F] \;\bigm|\; \gamma_\ell(f') = \gamma_\ell(f) \,\bigr\}
\end{aligned}
\]

\noindent\textbf{Approach 2: Hierarchical linkage-based steering.}\;
Seeds $\mathcal{F}'_\ell$ are defined in Step 4. Ward linkage is applied to
\[
A'_\ell \in \mathbb{R}^{F \times |T^+|}.
\]
For each seed feature $f$, the clustering function $\psi$ extracts features from the linkage subtree that contains $f$. The function add the leaves of the subtree:
\[
\psi_{\mathrm{link}}(A'_\ell, f)
=
\{\, f' \in \mathcal{T}_\ell(f) \,\}.
\]
During linkage-traversal at each step, $f'$ is only considered if the combined cluster size is less than 50 to retain strong association among features. This constrained approach may lead to empty feature subgroups.

\noindent\textbf{Approach 3: Single-token-driven steering.}\;
This approach concentrates the seed signal on the single most concept-aligned token $t^{*}$. The seed set is now layer-specific as shown in Step 4. The clustering function applies hierarchical linkage restricted to the activation vector of $t^{*}$ alone:
\[
    \psi_{\mathrm{single}}(A'_\ell,\, f)
    =
    \psi_{\mathrm{link}}\!\left(A'_\ell,\, f\right),
    \qquad f \in \mathcal{F}'_\ell.
\]

\paragraph{Step 6: Steering and Evaluation.}
For each layer $\ell \in \mathcal{L}$, we amplify the features in $\Phi_\ell$ within the SAE activation latent space for $\mathcal{M}$ during inference. Let $r(x) \in [1, 5]$ be the harmfulness score of the baseline response (unsteered), and $r^{*}(x, \ell) \in [1, 5]$ be the harmfulness score of the steered response produced after steering at layer $\ell$. Scores were assigned by an LLM judge. A layer $\ell$ is said to exhibit a \textit{feature vulnerability} for prompt $x$ if:
\[
    r^{*}(x,\, \ell) > r(x)
\]
Our objective is to characterize the distribution of feature vulnerabilities across $\mathcal{L}$ and across the 14 harm categories in $\mathcal{D}$, and to compare how the three instantiations of $\psi$ differ in the layers and categories they expose.

% \begin{figure*}[h]
%     \centering
%     \includegraphics[width=1.0\linewidth]{latex/figures/g23.png}
%     \caption{Approach 1. This figure for Gemma-2-2B represents cluster-based steering with original prompts. It shows the particular harm categories and mid-later layers that are relatively more prone to feature steering.}
%     \label{g23}
% \end{figure*}

% \begin{figure*}[h]
%     \centering
%     \includegraphics[width=0.75\linewidth]{latex/figures/g22.png}
%     \caption{Approach 2. This figure for Gemma-2-2B represents hierarchical linkage-based steering with original prompts. It shows the particular harm categories and mid-later layers that are relatively more prone to feature steering.}
%     \label{g22}
% \end{figure*}
 
% \begin{figure*}[h]
%   \centering
%   \includegraphics[width=1.0\linewidth]{latex/figures/g24.png} \hfill
%   \caption{Approach 3. This figure for Gemma-2-2B represents single-token-driven feature steering with original prompts. Similar to the other two methods this heatmap also shows (mid-late) layers as more vulnerable to feature steering.}
%     \label{g24}
% \end{figure*}

%  \begin{figure*}[h]
%   \centering
%   \includegraphics[width=1.0\linewidth]{latex/figures/g21.png} \hfill
%   \caption {Figure demonstrating that the later layers are relatively more steerable than the early-mid layers in Gemma-2-2b on all original datapoints.}
%     \label{g21}
% \end{figure*}

% Put this near the top of page 4 source location
\begin{figure*}[!t]
    \centering
    \setlength{\abovecaptionskip}{2pt}
    \setlength{\belowcaptionskip}{-6pt}

    \includegraphics[width=0.78\linewidth]{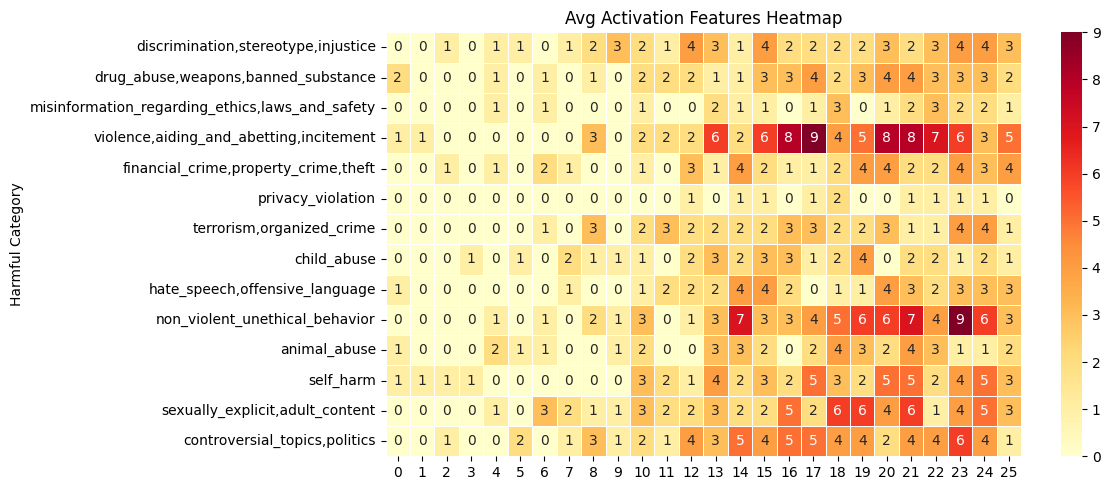}

    \includegraphics[width=0.58\linewidth]{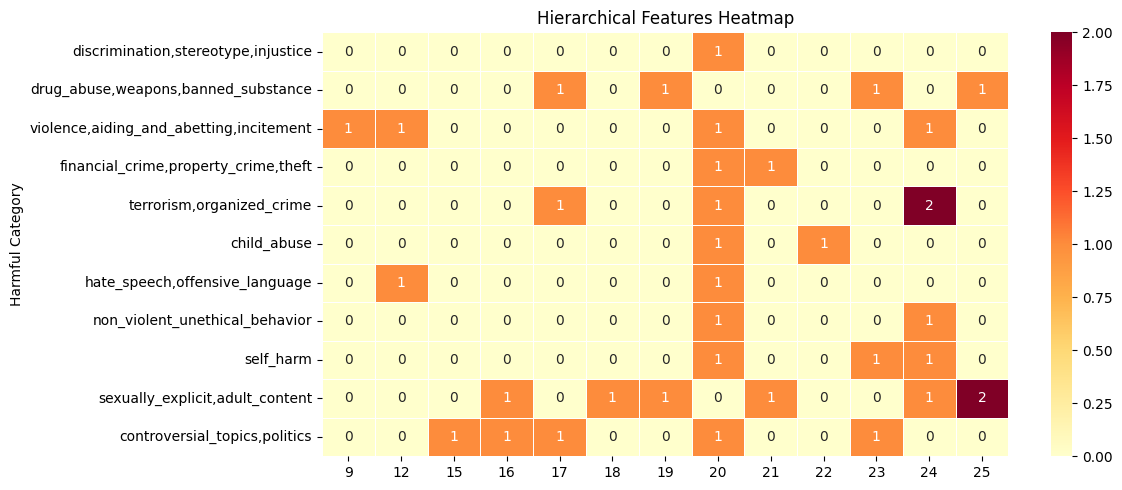}

    \includegraphics[width=0.78\linewidth]{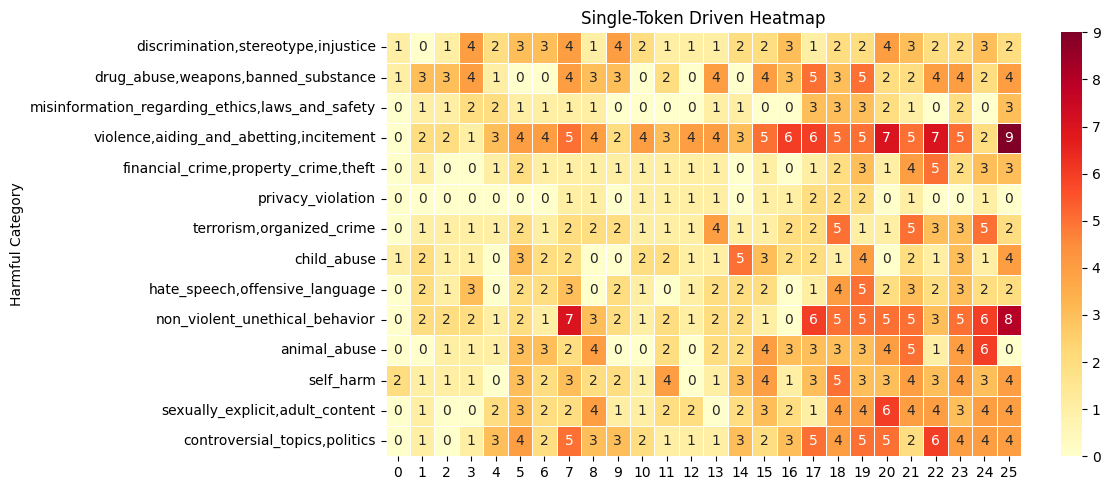}

    \includegraphics[width=0.78\linewidth]{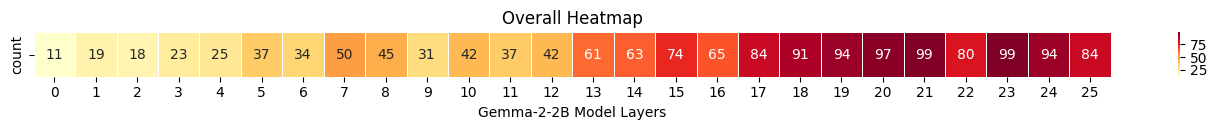}
    \vspace{-0.2em}

    \caption{Gemma-2-2B steering results with original prompts. From top to bottom: cluster-based, hierarchical-linkage-based, single-token-driven, and overall layer-wise vulnerable prompt counts. Mid-to-late layers show higher vulnerability across methods.}
    \label{fig:g2b_original_all}
\end{figure*}

\section{Experimental Setup}
\label{sec:exp_setup}

\paragraph{Models}

To extract the negative sentiment words/phrases from adversarial responses $z$, we employed Grok-4-1-fast-non-reasoning. We found that Gemma-2-2B produced repetitive tokens or continued generating within the context of the adversarial prompt when asked to perform this extraction. Grok-4-1-fast-non-reasoning, by contrast, reliably isolated concise negative-sentiment terms, likely due to its larger scale and broader pre-training data. We therefore use it as a preprocessing utility and as a judge model in our experiments; Gemma-2-2B remains the only model subject to feature analysis and steering.

We use Gemma-2-2B \citep{team2024gemma} as the target model for all steering experiments. For steering we used Neuronpedia's steering API\footnote{https://www.neuronpedia.org/api-doc\#tag/steering}. This choice is motivated by two factors. First, Gemma-2-2B is a lightweight, open-source model that permits full access to residual-stream activations across all 26 layers. Second, \citet{lieberum2024gemma} released Gemma Scope, that consists of an open-source suite of SAEs trained on every layer and sub-layer of Gemma-2-2B and other models. The availability of pre-trained SAEs allows us to extract 16k-dimensional feature activations without additional training overhead.

%  \begin{figure*}[h]
%   \centering
%   \includegraphics[width=1.0\linewidth]{latex/figures/g2b3.png} \hfill
%   \caption {Approach 1. This figure for Gemma-2-2B represents cluster-based steering with benign prompts. It shows the particular harm categories and mid-later layers that are relatively more prone to feature steering.}
%     \label{g2b3}
% \end{figure*}

%  \begin{figure*}[h]
%   \centering
%   \includegraphics[width=0.7\linewidth]{latex/figures/g2b2.png} \hfill
%   \caption {Approach 2. This figure for Gemma-2-2B represents hierarchical linkage-based steering with benign prompts. It shows the particular harm categories and mid-later layers that are relatively more prone to feature steering.}
%     \label{g2b2}
% \end{figure*}

%  \begin{figure*}[h]
%   \centering
%   \includegraphics[width=1.0\linewidth]{latex/figures/g2b4.png} \hfill
%   \caption {Approach 3. This figure for Gemma-2-2B represents single-token-driven feature steering with benign prompts. Similar to the other two methods this heatmap also shows (mid-late) layers as more vulnerable to feature steering.}
%     \label{g2b4}
% \end{figure*}

%  \begin{figure*}[h]
%   \centering
%   \includegraphics[width=1.0\linewidth]{latex/figures/g2b1.png} \hfill
%   \caption {Figure demonstrating that the later layers are relatively more steerable than the early-mid layers in Gemma-2-2b on all benign datapoints.}
%     \label{g2b1}
% \end{figure*}

\begin{figure*}[!t]
    \centering
    \setlength{\abovecaptionskip}{2pt}
    \setlength{\belowcaptionskip}{-6pt}

    \includegraphics[width=0.78\linewidth]{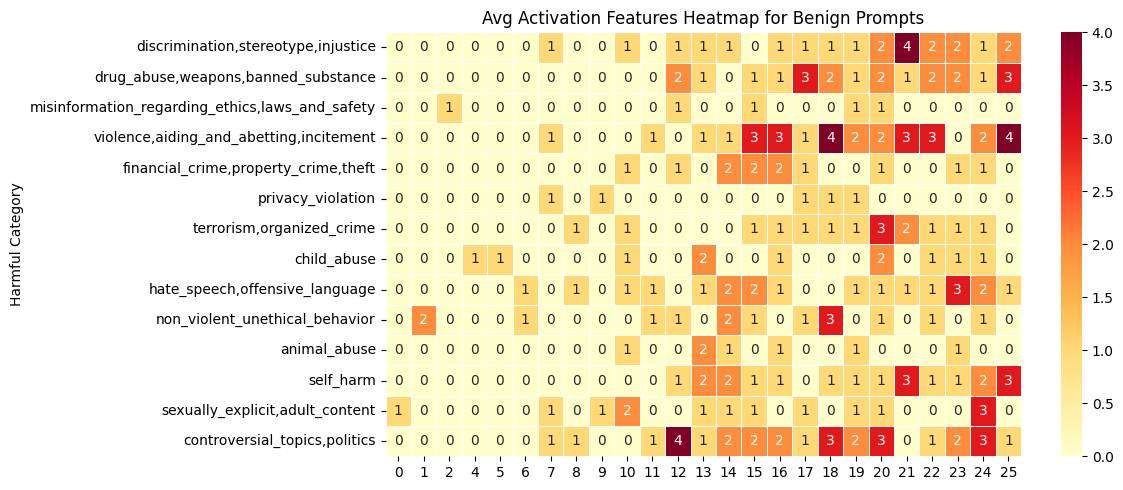}

    \includegraphics[width=0.58\linewidth]{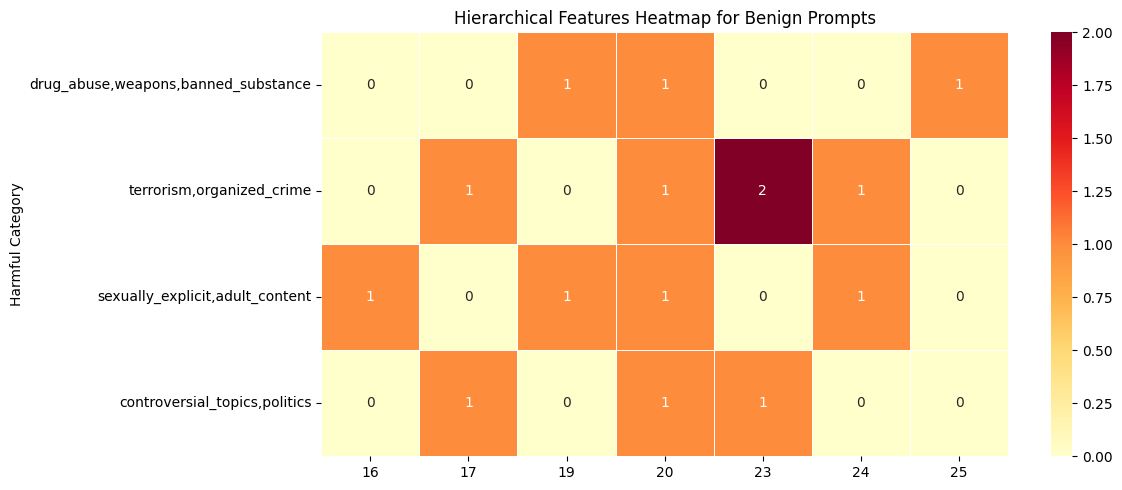}

    \includegraphics[width=0.78\linewidth]{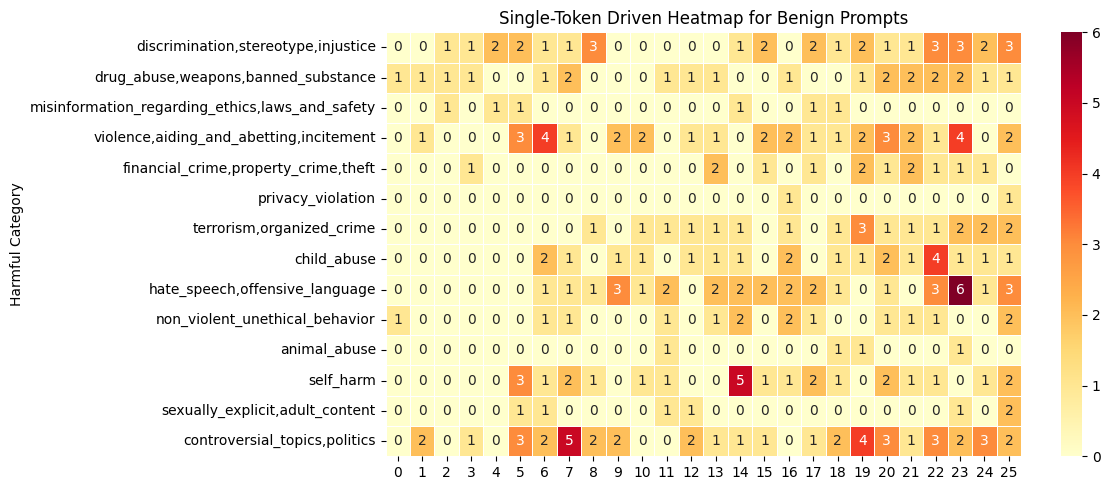}

    \includegraphics[width=0.78\linewidth]{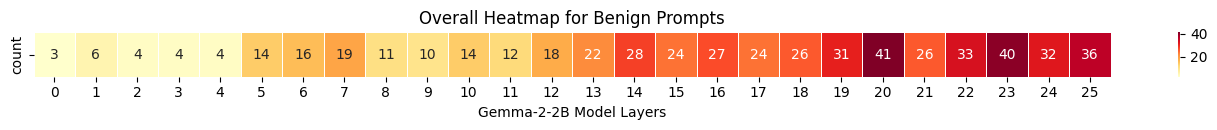}

    \caption{Gemma-2-2B steering results with benign prompts. From top to bottom: cluster-based, hierarchical-linkage-based, single-token-driven, and overall layer-wise vulnerable prompt counts. Steering vulnerability decreases compared with original prompts, but remains nonzero.}
    \label{fig:g2b_benign_all}
\end{figure*}

\paragraph{Dataset.}
We used the BeaverTails dataset \citep{ji2023beavertails}, a widely adopted benchmark for safety-alignment research. Each datapoint contains an adversarial prompt, a model response, a set of 14 harmful-category labels, and a binary safety flag (\texttt{is\_safe}). From this corpus, we sample 20 prompts per category. We limited the analysis to 20 prompts per unsafe category because mechanistic interpretability experiments are computationally expensive and time-intensive. Unlike standard prompting-based evaluation, each example requires extracting and analyzing internal model representations across layers, which makes large-scale analysis impractical. For categories with fewer than 20 eligible examples, we included all available examples, resulting in 265 prompts in total. We also kept only prompts that belonged to a single unsafe category. This was important because our goal was to connect the negative-sentiment words and phrases in each model response to a specific harm type. Including prompts that overlapped across multiple unsafe categories would make it difficult to determine which category was responsible for the observed features, introducing confounding into the analysis. We also test the model steering under controlled settings with the same features but with benign prompts. The only difference is that when steering the features, we do not use the original prompt; instead, we use a benign-looking version of it. For each original prompt, a benign version was created using Grok.

\paragraph{Evaluation Protocol and Baselines.}
We define the \emph{baseline} as the unsteered response that Gemma-2-2B generates for each adversarial prompt under default decoding. To quantify the effect of steering, we score both the baseline and steered response on a 1--5 harmfulness scale, where 1 is the safest response, and 5 denotes the most harmful. Scoring is performed by an LLM judge using the template introduced by \citet{qi2024finetuning}, which evaluates whether a response complies with a harmful request and the degree of detail it provides.

% To complement our feature-steering experiments with a suffix-based adversarial attack, we adapt the population-level GCG framework introduced by \citet{li-etal-2026-evaluating}, who propose a generalized GCG algorithm that searches for adversarial input perturbations by optimizing over SAE latent activation objectives rather than model outputs directly. We instantiate this framework in the \emph{untargeted, population-level, suffix} configuration: the attack appends a 3-token adversarial suffix to each prompt and minimizes the concept overlap between the perturbed and original SAE feature distributions, without steering toward any fixed target sequence. For Gemma-2-2B, we target layer 16 using the \texttt{gemma-scope-2b-pt-res} SAE release. For Gemma-2-9B-it, we similarly target layer 16 using the \texttt{gemma-scope-9b-pt-res} release. In both cases, optimization runs for 20 iterations with $m=300$ candidate tokens per position, top-$k=170$ SAE features, and a batch size selected adaptively from $\{100, 90, \dots, 10\}$ to accommodate GPU memory constraints. We evaluate this setup across all 14 BeaverTails harm categories.
To complement feature steering, we adapt the population-level GCG framework of \citet{li-etal-2026-evaluating}. We use the \emph{untargeted, population-level, suffix} setting, appending a 3-token suffix and optimizing SAE latent objectives rather than model outputs. For Gemma-2-2B, we target layer 16 using the corresponding Gemma Scope residual SAE, run 20 iterations with $m=300$ candidate tokens and top-$k=170$ SAE features, and evaluate across all 14 BeaverTails harm categories.
\section{Results and Analysis}
\label{res}

%We evaluate the three feature-grouping strategies of Section~\ref{prb_frm} across the 14 harm categories of BeaverTails and all 26 layers of Gemma-2-2B, under both the original adversarial prompts and their benign rewrites. Throughout, a layer exhibits a \emph{feature vulnerability} for a prompt when the steered response receives a higher harmfulness score than the unsteered default. 

The analysis is organized around the three questions that motivate the study: whether the grouping rule changes which features are recovered, whether the resulting vulnerability is shared across harm categories or specific to a few, and whether the harm-responsible features are properties of the prompt or transferable directions in the model.

\paragraph{A single harmful token is sufficient to localize harm, performing comparably to using the full set of harmful tokens.}
Cluster-based steering and single-token-driven steering are the two most effective strategies for increasing harmfulness, and both exceed hierarchical-linkage steering (Figure~\ref{fig:g2b_original_all}), particularly in the mid-to-late layers. Cluster-based steering becomes reliably effective from roughly layer 13 onward, whereas single-token-driven steering produces a broader pattern that reaches some early layers as well as the mid-to-late band. The two strategies differ in how they form the seed set: cluster-based steering considers all harm-aligned tokens, whereas single-token-driven steering depends entirely on the single most harm-aligned token. That they reach comparable effectiveness is the central result of this work. \textit{It shows that the grouping rule left unexamined in prior feature-clustering work is consequential, and that anchoring the search on one harmful token can recover a harm-responsible subgroup equal in effectiveness to the one that would otherwise depend on the full token set.} Harmful behavior is therefore carried not by an isolated feature but by a subgroup of co-activating features, and a single prompt token is a sufficient entry point for finding it.

Hierarchical-linkage steering is the most selective of the three strategies by construction. We treat two clusters as associated only when their merged cluster contains at most 50 members. Linkage nodes indexed above 16{,}383 can aggregate more than 50 features and would otherwise admit loosely connected directions. This constraint keeps the recovered subgroups tight, but it also excludes many features at a given layer; consequently, fewer prompts are steerable than under the other two strategies. This selectivity reflects the thresholding rule, not evidence that harm-responsible features are scarce.

\paragraph{The vulnerable layers are shared across harm categories, not confined to one.}
Our results make a different point from a purely global account of refusal \cite{arditi2024refusal, joad2026there}. Alignment failure is not localized to one harm category or to one global refusal axis. The main pattern is that steerability increases across several harm categories in the mid-to-late layers rather than appearing in only one category. At the category level, \texttt{violence},\texttt{aiding\_and\_abetting},\texttt{incit}\\\texttt{-ement} and \texttt{non\_violent}\texttt{\_unethical\_be}\\\texttt{-havior} are the most consistently steerable under both cluster-based and single-token-driven steering (Figure~\ref{fig:g2b_original_all}). Successful steering is concentrated in layers~14--25. For \texttt{violence},\texttt{aiding\_and\_abetting},\texttt{incite}\\\texttt{-ment}, both strategies peak at nine steered prompts. Single-token-driven steering also reveals one earlier effect, where \texttt{non\_violent}\texttt{\_unethical\_behavior} peaks sharply at layer~7, decreases between layers~8 and~16, and rises again from layer~17 onward. Apart from this early category-specific peak, the broader result is that vulnerable layers are shared across harm categories. Amplifying harm-responsible feature subgroups is sufficient to move the model from refusal to compliance, even when the steering begins from a single harmful token. This qualifies single-direction accounts of refusal \citep{arditi2024refusal}, because the relevant vulnerability is not only whether a global refusal vector is present or absent. It also depends on whether harm-responsible feature subgroups can be amplified in shared mid-to-late layers. Hierarchical-linkage steering shows a different pattern, with fewer steerable prompts spread across layers and a slight concentration in the final two layers, 24 and~25.

\paragraph{The harm-responsible features are largely prompt-specific.}
When the same feature subgroups are applied to benign rewrites of the prompts, vulnerability falls substantially but does not vanish (Figure~\ref{fig:g2b_benign_all}). \textit{This benign control separates features that are specific to the harmful prompt from those that transfer: the residual effect indicates that the two prompt versions share some semantic structure, but the features that actively fire are predominantly tied to the harmful prompt.} A subgroup can occasionally carry over to a benign prompt, yet the effect is weaker, which is consistent with the limited reach of hierarchical-linkage steering and points to a single design conclusion. Steering should target the top-activating features that co-activate with other features within a layer. Cluster-based steering remains the stronger of the two effective strategies on benign prompts and continues to track single-token-driven steering closely, while hierarchical-linkage steering shows only a slight rise at layer 23.

\paragraph{Where steering succeeds and where it does not.}
The three strategies are weakest in three regimes: benign prompts, early layers, and the restrictive hierarchical-linkage setting. This failure pattern clarifies that steering is not simply generating harmful behavior from any input. Instead, successful steering depends on a match between the prompt and the targeted feature subgroup. When the original prompt contains harm-relevant semantics, amplifying the corresponding subgroup can move the model from refusal to compliance. When the prompt is benign, the same intervention is much less effective because the input does not provide the semantic structure needed for that subgroup to become behaviorally active. This makes the result distinct from prompt generation or generic activation perturbation. The intervention does not create a new harmful request; it reveals when an existing harmful request has activated internal features that can be amplified. Early layers are also less steerable because they mainly encode lower-level lexical and syntactic information, whereas mid-to-late layers encode more task-relevant semantic features that are more susceptible to targeted amplification. Hierarchical-linkage steering is weaker for a different reason. Its strict cluster-size rule keeps recovered subgroups tight, but it reduces coverage by excluding many features at a given layer. These failures show that harmful behavior emerges from interactions among prompt semantics, layer position, and subgroup coverage, rather than from the feature subgroup alone.

\paragraph{Quantitative and Qualitative Assessment.}
We performed more than 10{,}000 steering evaluations in total. For the original adversarial prompts, $17.4\%$ of steered responses (1{,}798 of 10{,}306) received a higher harmfulness score than their unsteered default, against only $6.0\%$ (618 of 10{,}307) for the benign rewrites, a near-threefold gap that quantifies the prompt-specificity reported above. The strongest evidence that the recovered subgroups are causally sufficient comes from responses that began as non-harmful content (default score 1): $3.70\%$ were driven to maximal harm (score 5) and a further $1.10\%$ to score 4, so $4.8\%$ of non-harmful content were overturned outright by amplifying a harm-responsible subgroup. These overturned safety compliance cases are the cases in which feature amplification, rather than any change to the prompt, is solely responsible for the shift from safe to unsafe.

We manually studied a few feature subgroups for layer 23 that show the highest vulnerability in the overall analysis in Figure~\ref{fig:g2b_original_all}. We observed in the case of cluster-based steering, for a prompt from \texttt{self\_harm} category, interpretable features `descriptions of physical appearances and qualities', `terms related to urban development and urbanism', and `elements related to betrayal and conflict between characters' when amplified, changed the score from 1 to 5. In case of \texttt{financial\_crime,property\_crime,th}\\\texttt{-eft} at layer 23, we observed the three interpretable features: `references to weather conditions and forecasts', `code and programming constructs related to OpenGL specifications and functions', `mathematical symbols and expressions related to scales and measurements' served highly effectively in steering. Among these most of the interpretable features show a link to the harmful category.

\begin{table}[t]
\centering
\small
\setlength{\tabcolsep}{3pt}
\renewcommand{\arraystretch}{0.92}
\begin{tabularx}{\columnwidth}{@{}Xcc@{}}
\toprule
\textbf{Category} & \multicolumn{2}{c}{\textbf{Count}} \\
\midrule
\textit{Total (All Categories)} & \textit{26} & \textit{17} \\
Animal Abuse & 1 & 2 \\
Child Abuse & 0 & 2 \\
Controversial Topics / Politics & 1 & 1 \\
Discrimination / Stereotypes / Injustice & 1 & 1 \\
Drug Abuse / Weapons / Banned Substances & 4 & 2 \\
Financial Crime / Property Crime / Theft & 3 & 0 \\
Hate Speech / Offensive Language & 1 & 0 \\
Misinformation (Ethics, Laws, Safety) & 1 & 0 \\
Non Violent Unethical Behavior & 0 & 0 \\
Privacy Violation & 3 & 0 \\
Self Harm & 3 & 1 \\
Sexually Explicit / Adult Content & 4 & 2 \\
Terrorism / Organized Crime & 2 & 1 \\
Violence / Aiding \& Abetting / Incitement & 2 & 5 \\
\bottomrule
\end{tabularx}
\caption{Distribution of Harmfulness Score 5 in Gemma 2-2B by Category, across two approaches: fixed-layer and clustering.}
\label{tab:gemma2b_score_distribution}
\end{table}

\paragraph{Baseline comparison.}
We use a fixed-layer baseline to establish the level of severe harmfulness produced by an unconstrained intervention. The baseline is applied at layer~16 to all 265 prompts and produces 26 responses with a harmfulness score of 5. This baseline is relevant because it measures full-coverage attackability. It asks how often severe harmful behavior can be induced when every prompt is perturbed at the same layer. Our method asks a different question. It measures whether severe harmfulness remains inducible when steering is restricted to recoverable harm-associated feature subgroups.

This distinction is important because the two columns in Table~\ref{tab:gemma2b_score_distribution} have different denominators. The baseline count is computed over all 265 prompts. Our count is computed only over prompts that satisfy the eligibility constraints of the corresponding steering strategy, since many prompts are filtered out when no valid subgroup is recovered. Therefore, the value of 17 should not be read as a direct raw-count loss against 26. It shows that even under a stricter, lower-coverage intervention regime, localized feature-subgroup steering still recovers a substantial number of score-5 responses.

The category-level results further show why this comparison is informative. In Table~\ref{tab:gemma2b_score_distribution}, each value for our method reports the maximum score-5 count obtained across the three steering strategies. The category \texttt{violence, aiding\_and\_abetting, incitement} retains all 20 prompts under both cluster-based and single-token-driven steering. In this category, our method produces five score-5 responses, compared with two under the baseline. Our method also exceeds the baseline for \texttt{animal\_abuse} and \texttt{child\_abuse}, and matches the baseline for \texttt{controversial\_topics,politics} and \texttt{discrimination, stereotype, injustice}. These cases show that the proposed method does not merely reproduce the baseline distribution. It exposes category-specific vulnerabilities that a fixed-layer baseline does not fully capture. The aggregate comparison should therefore be interpreted as a trade-off between coverage and mechanistic selectivity. The baseline has full coverage but limited mechanistic specificity. Our method has lower coverage by design, but it identifies cases where severe harmful behavior can be induced through sparse, localized, and recoverable feature subgroups. This supports the central claim that harmfulness is not only a product of broad fixed-layer perturbation. It can also be carried by co-activating feature subgroups whose amplification is sufficient to move the model from refusal to compliance.

\paragraph{Additional Results.}

\begin{figure*}[!t]
    \centering
    \setlength{\abovecaptionskip}{2pt}
    \setlength{\belowcaptionskip}{-6pt}

    \textbf{Original Prompts}\hspace{0.34\linewidth}\textbf{Benign Prompts}

    \vspace{0.3em}

    % Approach 1
    \includegraphics[width=0.42\linewidth]{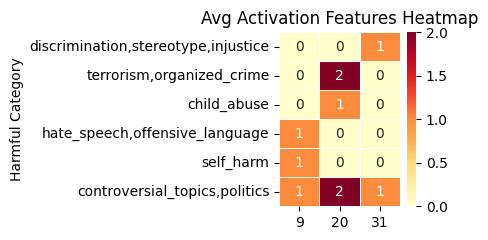}
    \includegraphics[width=0.52\linewidth]{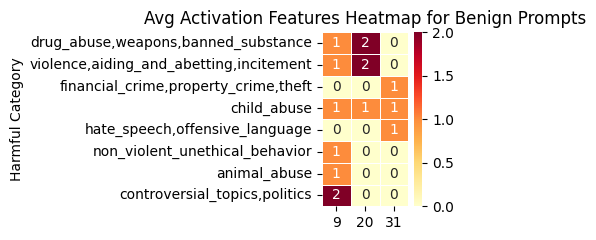}

    \vspace{0.55em}

    % Approach 2
    \includegraphics[width=0.38\linewidth]{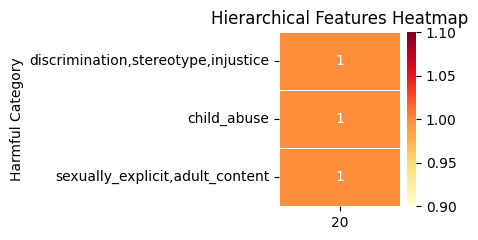}
    \hspace{0.01\linewidth}
    \includegraphics[width=0.48\linewidth]{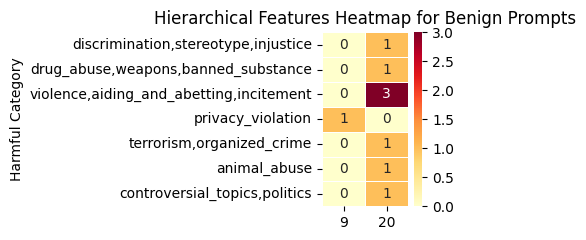}

    \vspace{0.55em}

    % Approach 3
    \includegraphics[width=0.42\linewidth]{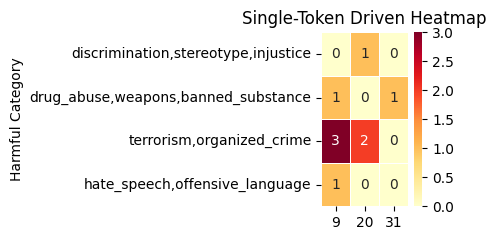}
    \hspace{0.01\linewidth}
    \includegraphics[width=0.48\linewidth]{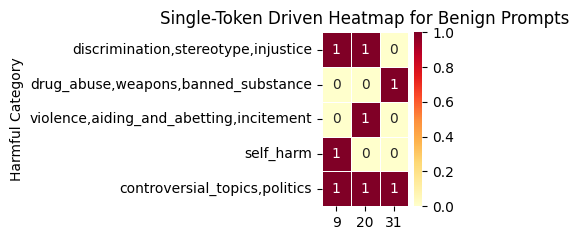}

    \vspace{0.55em}

    % Overall layer-wise count
    \includegraphics[width=0.3\linewidth]{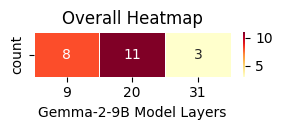}
    \hspace{0.01\linewidth}
    \includegraphics[width=0.3\linewidth]{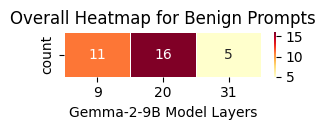}

    \vspace{-0.3em}

    \caption{Gemma-2-9B steering results for original and benign prompts. Rows correspond to cluster-based steering, hierarchical-linkage-based steering, single-token-driven feature steering, and overall layer-wise vulnerable prompt counts.}
    \label{fig:g9b_original_benign_all}
\end{figure*}

We also tested with Gemma-2-9B features. However, as Neuronpedia doesn't support steering of Gemma-2-9B, we performed steering experiments on Gemma-2-9B-IT with features identified from layers 9, 20, and 31 of Gemma-2-9B. This shouldn't make a difference as \citet{lieberum2024gemma} mentioned that base SAEs transfer well to IT SAEs. The authors made it clear in \href{https://huggingface.co/google/gemma-scope-9b-it-res}{Gemma Scope 9B IT residual SAEs} that the IT SAEs offer only a marginal improvement and PT SAEs are sufficient for the corresponding IT model, such as using Gemma 2 9B PT SAEs to interpret Gemma 2 9B IT. Therefore, we target Gemma-2-9B-IT using SAE features derived from Gemma-2-9B. Below, we detail our results on Gemma-2-9B-IT and present our findings as demonstrated in Figure~\ref{fig:g9b_original_benign_all}:

\begin{itemize}[leftmargin=*, noitemsep]
\item Cluster-based steering results with original prompts in Gemma-2-9B-IT show slightly higher vulnerability in layer 20, with 2 prompts being the highest. The benign prompts show slightly higher vulnerability in layers 9 and 20. Steering remains limited.

\item Hierarchical linkage-based steering shows that for original prompts, the identified feature subgroups have zero to limited steering effects on Gemma-2-9B-IT. However, benign prompts demonstrate a relatively increased vulnerability, with 3 being the highest number of prompts across categories in layer 20. Only in this case and cluster-based steering, features slightly transfer well to benign prompts.

\item Single-token-driven feature steering with original prompts show that early layer 9 is more vulnerable than layer 20. This is consistent with Gemma-2-2B findings. Benign prompts, again indicate reduced vulnerability. However, the vulnerability pattern is broader and is distributed across the three layers.

\item Cross-approach summary indicate overall vulnerability across all approaches and categories. With original prompts, layers 9 and 20 showed increased steerability than layer 31. Benign prompts also show a similar pattern. In case of Gemma-2-9B-IT majorly early to mid layers show improved steerability than later layers.

\end{itemize}

The subspace generators used in this research were introduced by \citet{wu2025axbench} and are available at 
\href{https://huggingface.co/pyvene/gemma-reft-2b-it-res-generator}{Gemma-ReFT-2B-IT} and 
\href{https://huggingface.co/pyvene/gemma-reft-9b-it-res-generator}{Gemma-ReFT-9B-IT}. Motivated by similar considerations as in our use of SAEs, we used instruction-tuned subspaces as the base model subspaces.

\section{Conclusion}
\label{conclu}

In this research, we aim to identify subgroups of features that, when steered, transition a model towards unsafe behavior. Results show that finding such features is plausible given the three feature-grouping strategies we employed. Additionally, a single prompt-specific negative concept-aligned token is enough to manipulate the model towards harmfulness. We also noticed that not only the mid-late layers of the target model are more subject to steering but in few cases early layers might also be highly vulnerable. Features identified for original prompts do transfer to benign prompts but are  limited and majorly prompt-specific.

\section*{Limitations}

In this research, we used SAEs to extract feature activations from model layers. We used already available SAEs, as training them is both resource and time intensive. Our study is limited to open-source models only. At this point we tried with Gemma-2-2B and Gemma-2-9B-IT, however, the proposed methodology can be adapted for other open-source models as well. This study was performed to understand the internal mechanics of the model during adversarial pressure. Thus, it specifically aims for a mechanistic understanding of the adversarial task and may not be generalizable to other tasks or domains. We used Grok-4-1-fast-non-reasoning as a judge model, thus, there may exist bias towards either giving a higher score or a lower score to the responses. Ground-truth severity labels for harmful responses are unavailable.

% \bibliography{custom}

\bibliography{custom}

\newpage
\appendix
\section{Appendices} 
\label{sec:appendix}

% \subsection{Implementation Details}

% We performed our experiments on L40s or rtx\_8000 GPUs depending on the availability of resources. We utilized approximately 500 GPU hours for our experiments. Open-source model like Gemma-2-2B can be easily downloaded from Hugging Face. The subspace generator $G$, introduced in Section \ref{prb_frm} is gemma-reft-2b-it-res-generator\footnote{https://huggingface.co/pyvene/gemma-reft-2b-it-res-generator}. We used open-source Python libraries in our research.

% For population-based suffix attacks, we used the following hyperparameters: (layer\_num=16), (suffix\_len=3), (num\_iters=20), (m=300), and (k=170), with mode=suffix and level=population. To handle memory constraints, batch size was selected from $\{100, 90, 80, 70, 60, 50, 40, 30, 20, 10\}$ using a fallback strategy. We evaluated up to 999 sample indices per dataset and terminated early when the dataset boundary was reached. Experiments were run over 14 safety-related dataset categories.

% For inference, we loaded models with device\_map=``auto'' and torch\_dtype=bfloat16. Responses were generated greedily (do\_sample=False) with (max\_new\_tokens=150). For automatic evaluation, we used a judge model (grok-4-1-fast-non-reasoning) to assign harmfulness scores on a 1--5 scale, with up to 3 retry attempts per sample (exponential backoff) and incremental CSV logging.

\subsection{Ethical Considerations}

This research was conducted to obtain a mechanistic understanding of the model's behavior on an adversarial task. The ultimate goal is to help the broader research community, especially AI safety experts, develop more robust and safer models over time. This research was not undertaken to build a product and commercialize it in any way. Nor has it involved humans till now. This was a black-box analysis in which we gained an understanding of the model's behavior when specific features were steered. We used the BeaverTails dataset, which is publicly available under the CC BY-NC-4.0 license.

\subsection{Related Work}

\paragraph{From input-output attacks to internal mechanisms.}
Adversarial attacks established that safety alignment is brittle. Simple jailbreak prompts that use role-play, competing objectives, or refusal suppression bypass guardrails in aligned models \citep{wei2023jailbroken}. Automated methods made this systematic: Greedy Coordinate Gradient (GCG) optimizes adversarial suffixes through token-level gradients \citep{zou2023universal}, Tree of Attacks with Pruning (TAP) refines prompts through black-box branching and pruning \citep{mehrotra2024tree}, and genetic and reinforcement-learning approaches reach high attack success on open-source models \citep{lapid2023open}. The limited transfer of these attacks to proprietary models indicates that success depends on model-specific internal properties rather than universal prompt structures \citep{lin2025understanding}, which motivates looking inside the model. Interpretability has begun to do so, but in a safe direction. \citet{arditi2024refusal} shows that refusal is mediated by a single direction in the residual stream, and \citet{obrien2024steering} amplifies sparse autoencoder (SAE) features that mediate refusal to harden a model at inference time. Both localize safety to a single direction or feature, and both steer toward refusal. This leaves open whether harmful behavior is carried by one feature or by a group of interpretable features acting together, and whether the harmful tokens in the prompt can themselves drive the search for those features.

\paragraph{Decomposing activations and grouping features.}
Mechanistic interpretability reverse-engineers networks into human-understandable computation, from individual circuits such as induction heads and curve detectors \citep{olah2018building, cammarata2020curve, nanda2023progress} to features distributed in superposition, where neurons encode many unrelated concepts \citep{elhage2021superposition}. SAEs address this by decomposing activations into sparse, monosemantic features \citep{cunningham2023sparse}, which gives a tractable unit for analysis and steering. Recent work treats single features as the unit of safety, but the assumption that a behavior maps to one feature is increasingly questioned: abstract concepts such as refusal appear distributed across several features rather than isolated in one \citep{yeon2025gsae}. Work that groups SAE features rather than treating them singly is scarce, and what exists serves a different end. \citet{gong2025signal} apply agglomerative clustering to features in Pythia-70M and GPT-2-Small, but to study cross-model interference between semantically unrelated pairs, not to localize a behavior. The grouping rule itself has gone unexamined. No prior work asks whether anchoring the search on the single most harm-aligned token recovers the same subgroup that aggregating over all harm-aligned tokens would.

\paragraph{Causal attribution and the direction of steering.}
Establishing that a feature is responsible for a behavior, rather than merely correlated with it, requires intervention. Cross-layer transcoders replace MLPs with interpretable features and enable feature-feature interaction graphs \citep{anthropic2025circuit}, and attribution graphs trace which features causally contribute to an output \citep{anthropic2025attribution}, but these analyses target benign tasks such as poem generation. Where SAEs meet safety, the steering runs one way. \citet{obrien2024steering} amplify refusal features to resist jailbreaks, GSAE assembles safety-relevant directions to enforce adaptive refusals \citep{yeon2025gsae}, and CC-Delta selects jailbreak-relevant features to mitigate attacks \citep{assogba2026sparse}. Each amplifies features that make the model safer. None amplifies harm as a causal probe to test where the vulnerability lives. Whether amplifying harm-responsible subgroups is sufficient to break alignment, which layers carry that vulnerability, and whether the pattern is shared or category-specific across harm types, remain open. A benign-prompt control is also needed to separate features that are specific to the harmful prompt from those that transfer.

\end{document}